\documentclass[a4paper,fleqn]{cas-dc}

\usepackage[numbers]{natbib}

\usepackage[justification=centering, singlelinecheck=false]{caption}
\usepackage{subcaption}
\usepackage{stfloats}
\usepackage{placeins}
\usepackage{float}
\usepackage{needspace}
\usepackage{tabularx}
\usepackage{booktabs}    % 你已在用
\usepackage{makecell}    % 处理长单元格或多行内容（可选）

\usepackage{hyphenat} % 断字支持
\usepackage{seqsplit} % 长单词分割
\usepackage{microtype} % 微调排版
\usepackage{ragged2e} % 更好的对齐控制
\usepackage[breakwords]{xurl} % URL自动换行

\usepackage{amssymb}
\usepackage{amsmath}
\usepackage{booktabs}
\usepackage{enumitem}
\usepackage{color}

\usepackage{afterpage}
\def\tsc#1{\csdef{#1}{\textsc{\lowercase{#1}}\xspace}}
\tsc{WGM}
\tsc{QE}
\begin{document}
	\let\WriteBookmarks\relax
	\def\floatpagepagefraction{1}
	\def\textpagefraction{.001}
	\let\printorcid\relax % 可去掉页面下方的ORCID(s)

	% \shorttitle{<short title of the paper for running head>} 
	\shorttitle{A Lightweight Convolution and Vision Transformer integrated model with Multi-scale Self-attention Mechanism}    
	
	% \shortauthors{<short author list for running head>}
	\shortauthors{Yi Zhang et al.}
	
	\title[mode = title]{A Lightweight Convolution and Vision Transformer integrated model with Multi-scale Self-attention Mechanism}  
	
	%% Title footnote mark
	%% eg: \tnotemark[1]
	% %\tnotemark[<tnote number>] 
	%\tnotemark[1,2]
	\tnotemark[1]
	\tnotetext[1]{This research was funded by the Intelligent Policing Key Laboratory of Sichuan Province, No.ZNJW2024KFMS004.}

	\author[1]{Yi Zhang} 
	
	\author[1]{Lingxiao Wei}
	
	\author[1]{Bowei Zhang}
	
	\author[2]{Ziwei Liu}
	
	\author[3]{Kai Yi}
	\cormark[1]
	
	\author[1]{Shu Hu}
	
	\address[1]{Department of Computer Science, Sichuan University, China}
	\address[2]{School of Computer Science, Nanyang Technological University, Singapore}
	\address[3]{Sichuan Police College,Intelligent Policing Key Laboratory of Sichuan Province, Luzhou, China}
	
	\cortext[1]{\parbox[t]{\linewidth}{%
			Corresponding author.\\
			E-mail address: yikai@scpolicec.edu.cn (K. Yi).%
	}}

	\begin{abstract}
		Vision Transformer (ViT) has prevailed in computer vision tasks due to its strong long-range dependency modelling ability. \textcolor{blue}{However, its large model size and weak local feature modeling ability hinder its application in real scenarios. To balance computation efficiency and performance in downstream vision tasks, we propose an efficient ViT model with sparse attention (dubbed SAEViT) and convolution blocks. Specifically, a Sparsely Aggregated Attention (SAA) module has been proposed to perform adaptive sparse sampling and recover the feature map via deconvolution operation,} which significantly reduces the computational complexity of attention operations. In addition, a Channel-Interactive Feed-Forward Network (CIFFN) layer is developed to enhance inter-channel information exchange through feature decomposition and redistribution, which mitigates the redundancy in traditional feed-forward networks (FFN). Finally, a hierarchical pyramid structure with embedded depth-wise separable convolutional blocks (DWSConv) is devised to further strengthen convolutional features. Extensive experiments on mainstream datasets show that SAEViT achieves Top-1 accuracies of 76.3\% and 79.6\% on the ImageNet-1K classification task with only 0.8 GFLOPs and 1.3 GFLOPs, respectively, demonstrating a lightweight solution for fundamental vision tasks. 
	\end{abstract}
	
	
	%\item highlight-1
	%\item highlight-2
	%\item highlight-3
	
	% Keywords
	
	\maketitle
	
	% Main text
	\section{Introduction}
	
	ViT has achieved remarkable results in various vision tasks (e.g. classification\cite{wang2023visual}, detection\cite{10177966} and segmenta\-tion\cite{pmlr-v202-liang23h} etc.) due to its powerful global modeling capabilities. Existing methods (e.g.\cite{guo2022cmt}, \cite{yu2022metaformer}) tried to integrate convolution neural network (CNN) with prior knowledge and local feature extraction capability into ViT. To enhance model accuracy, Ren et al.\cite{ren2022shunted} refined the original attention computation by introducing multi-scale self-attention mechanisms. However, \textcolor{blue}{the additional multi-scale computation modules would cause significant computational overhead, which impede} their applications in real-time scenarios. To reduce computation cost, traditional methods employed multi-level hierarchical architectures that down-sample feature maps progressively or adopted sliding window strategies in spatial dimension. However, they relied on a fixed sliding window, which can easily ignore the correlation of local information. Other methods\cite{wang2021pyramid} performed sub-sampling on keys (K) and values (V) in attention computation to further reduce the computation cost. \textcolor{blue}{On the other hand, Wang et al.\cite{wang2022pvt} attempted to enhance the expressivity of FFN by appending convolutional blocks or DWSConv to improve local feature modeling, but it overlooked inter-channel interactions.}
	
	To address the above mentioned problems, \textcolor{blue}{we propose a simple yet effective backbone for multiple vision tasks, where we endow ViT with local perception and long-range modeling ability.} Firstly, we integrate convolutional feature processing modules into the model to incorporate convolutional prior knowledge. Then, we adopt a sparse attention approach for feature extraction to reduce the computational cost of attention operation. By leveraging the redundancy of similar pixels in the input image, the aggregation of local features and sparse sampling are performed to ensure long-range dependencies of attention. Finally, to alleviate potential accuracy loss caused by sparse attention and to enhance inter-channel interactions, we replace FFN layer by CIFFN, which decomposes feature channels and reassign feature weights explicitly to preserve representational capacity under lightweight constraints.
	
	Built upon the above design principle, we propose SAEViT, a lightweight model that combines the strengths of CNNs and ViT. SAEViT originates from convolution+Trans-former hybrid structure and sparse attention, while we focus on lightweight design for edge devices by striking a balance between performance and efficiency. Specifically, SAEViT’s sparse attention module achieves faster computation than Spatial Reduction Attention (SRA) and Windowbased Attention, while it does not introduce additional computational overhead. We evaluate model performance primarily through accuracy and throughput. Throughput measures inference efficiency, reflecting the real-time applicability and computational resource utilization. Higher throughput indicates more images/data samples processed per second. We develop 2 model variants with different computational complexities: Tiny (T) and Extra Small (XS), but both have $\leq10$M parameters and $\leq1.5$GFLOPs. \textcolor{blue}{Aiming at resource restricted situations, we compare our model with other lightweight models. It turns out that SAEViT achieves better performance, while it indeed has fewer number of parameters and lower computation cost.} Experiments include image classification, object detection, and semantic segmentation. Take classification for instance, SAEViT-t achieves 76.3\% Top-1 accuracy with 0.8 GFLOPs on ImageNet, while the SAEViT-xs variant reaches 79.6\% with 1.3 GFLOPs. Comparison of throughput and latency will be reported in Section \ref{throughput}. 
	
	In a nutshell, the main contributions of this work include:
	
	\begin{itemize}
		\item A novel ViT framework has been proposed by combining CNN with ViT. It possesses the feature prior capabilities of CNN as well as the long-range dependency modeling ability of attention mechanisms.  
		\item The traditional aggregation attention mechanism has been redesigned to address the high computational cost. Compared to traditional window attention and SRA, SAEViT achieves significantly higher throughput. 
		\item A channel-interactive Feed-Forward Network has been developed to counteract potential accuracy degradation caused by sparse attention. Meanwhile, channel interaction has been performed to improve the expressivity of FNN under lightweight constraints.
	\end{itemize}
	
	\section{Related Works }
	\subsection{Vision Transformer}
	The rise of attention mechanisms promoted the popularity of ViT in the field of computer vision. As a result, a series of excellent works have emerged \cite{liu2021swin, tu2022maxvit, hatamizadehfastervit}, which significantly improves the performances of ViT based models (in terms of efficiency and accuracy). To enhance the overall feature extraction and generalization capabilities of models, many works focus on optimizing ViT's computational paradigm, particularly through the fusion of convolution blocks and attention mechanisms. Extensive experiments demonstrate that integrating CNN with ViT improves feature extraction. Other approaches \cite{zhang2022nested, wang2021pyramid, liu2021swin} aim to improve feature utilization efficiency by facilitating cross-layer feature interactions. 
	
	Recently, huge efforts have been made in reducing the spatial dimensions of Multi-Head Self-Attention (MHSA) module, which is normally a computational bottleneck in ViT architectures. MHSA maps input tensors into key, query and value tensors. Recent studies revealed that key and value tensors can be subsampled with minimal accuracy loss, achieving a better efficiency-accuracy trade-off. For instance, MaxViT\cite{tu2022maxvit} employed blocked local attention and dilated global attention to perform global and local spatial interactions across arbitrary inputs. Biformer\cite{zhu2023biformer} introduced a novel bi-level routing attention mechanism, enabling flexible computational allocation and content-aware dynamic sparse attention for improved efficiency and performance. Clusterfomer\cite{NEURIPS2023_c9ef471a} utilized clustering mechanism that is integrated within Transformer architecture to realize a universal vision model. ProMotion\cite{10657486} is a Transformer based motion model that calculates optical flow and estimates scene depth. Considering the original intention of lightweight design, we perform sparse attention operation on the feature maps rather than sub-sampling keys and values. 
	
	\subsection{The Feed-Forward Network}
	
	The FFN layer is a critical component in ViT. Its primary role is to introduce non-linear transformations and enhance the model’s representational capacity. By expanding and compressing channel dimensions in high-dimensional spaces, it enriches the feature representation of individual tokens within the architecture. It affects both the model’s expressiveness and ViT’s computational efficiency. \textcolor{blue}{Subsequent works} attempted to enhance FFN layers through convolutional operations to strengthen local features (e.g. PVTv2\cite{wang2022pvt}, SRFormer\cite{zhou2023srformer}, LightViT\cite{huang2022lightvit}). Others, from architectural perspective, modified the residual connections by designing inverted residual blocks to enhance cross-layer gradient propagation (e.g. CMT\cite{guo2022cmt}). Partially inspired by the above works, we focus on designing learnable parameters of FFN that interact with input feature maps to dynamically redistribute channel-wise parameters. Combining with SAA, our approach effectively boosts feature extraction capabilities \textcolor{blue}{with relatively high computational efficiency.}
	
	\subsection{Lightweight ViT}
	Considering that ViT suffers high computational cost, which hinders its application in resource constrained scenarios. Designing lightweight ViT is a research area of great significance. Current lightweight ViT designs mainly focus on latency reduction on real devices. \textcolor{blue}{They optimized GPU and CPU throughputs to reduce the} computational overhead of attention mechanisms and overall architecture. RepViT\cite{wang2024repvit} replaced the Transformer’s multi-head self attention (MSA) module with lightweight convolutional blocks. It adopted a multi-branch structure during training to enrich feature representations, and merged these branches into an efficient convolutional layer for inference. Meanwhile, it adjusted the number of channels and network depth to reduce parameters, \textcolor{blue}{which realizes lower latency on edge device.} MobileViT\cite{mehta2021mobilevit} employed a hybrid architecture with unfold–transform–fold operations, achieving higher accuracy while reducing FLOPs. EfficientViT\cite{cai2023efficientvit} effectively boosted both accuracy and GPU throughput by revising the inverted bottleneck structure and introducing cascaded group attention. 
	
	\section{Method}
\begin{figure*}[htbp]
		\centering
		\includegraphics[width=\textwidth]{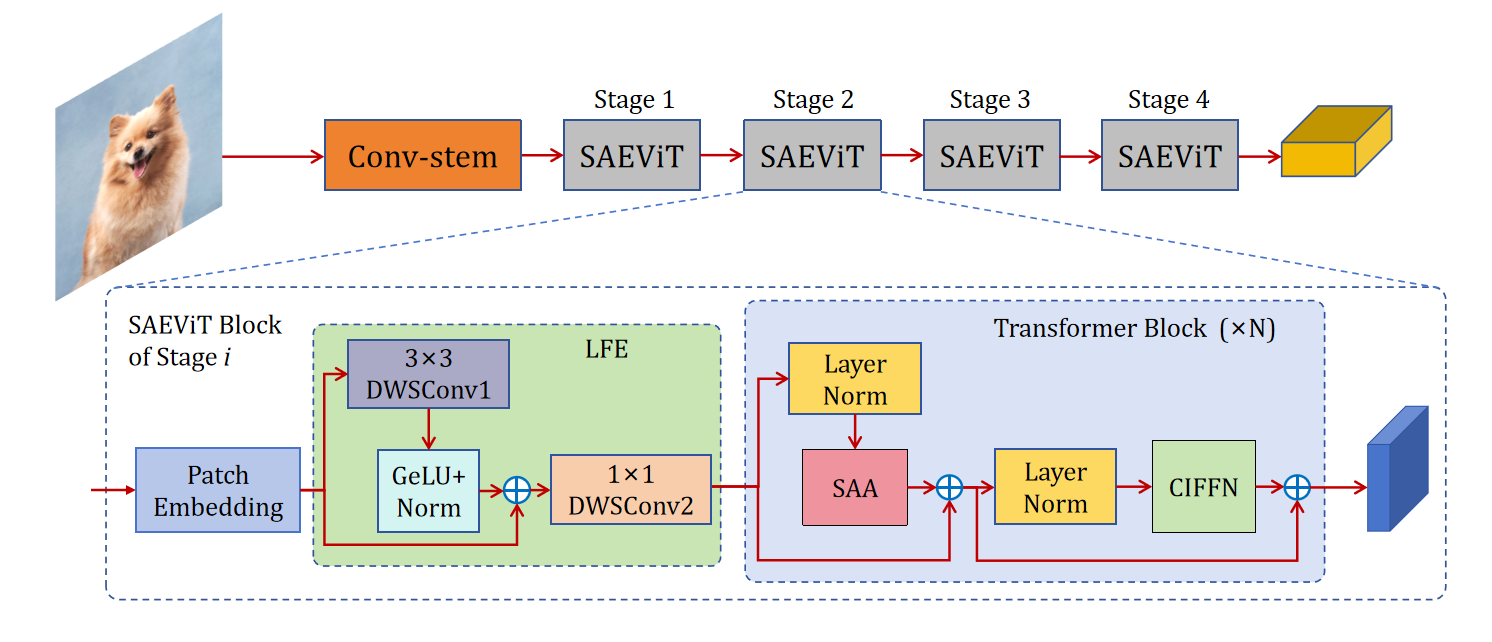}
		\caption{ 
			The overall architecture of the proposed SAEViT. The input is first processed by convolutional stems followed by 4 stages of SAEViT blocks, each of which applies Patch Embedding followed by Local Feature Extraction (LFE) and stacked Transformer modules with Sparsely Aggregated Attention (SAA) and Channel‐Interactive FFN (CIFFN).}
		\label{fig:fig1}
		\vspace{10pt}
	\end{figure*}

	Our goal is to combine the strength of convolutional neural networks (CNN) with the attention mechanisms, and develop a lightweight ViT based architecture. Following the design principle of pyramid network, our model is divided into 4 stages, each of which produces feature maps of different scales to capture multi-scale image information for detection and segmentation tasks. Between each stage, Patch Embedding layers (which consists of convolution and normalization operations) are adopted to down-sample the resolution to half of its original value. \textcolor{blue}{The effectiveness of the convolution stems in improving the accuracies in various vision tasks will be proved in Section \ref{backbone}.} Accordingly, our convolution stem is built upon 2 convolution layers (with kernel size=3, stride=1, padding=1). Meanwhile, we stack various Transformer blocks at each pyramid layer to meet diverse task requirements and computational resource constraints. Each Transformer blocks share the same internal structure, which is comprised of a SAA module and a CIFFN. The overall architecture of our proposed model is shown in Fig. \ref{fig:fig1}. 
	
	To construct a hybrid convolution-ViT structure, we connect Local Feature Extraction (LFE)\cite{zhang2025mscvit} module with modified Transformer block. LFE is an embedding layer with DWSConv and an activation function, which refines the input features to the Transformer block. Inside the Transformer block, SAA module rectifies the sampling strategy by performing the sparse block-wise sampling on the feature map in an adaptive way, which reduces the computational load of traditional attention mechanisms. To deal with information redundancy and limited feature expressivity in high-dimensional features, \textcolor{blue}{we employ inter-channel interaction mechanism in FFN.} To be specific, a 1×1 convolution is firstly applied to map the input features into a higher dimensional space. Then, depth-wise convolution (DWConv) is used to enhance features within each channel, enriching the feature content. An activation function is then introduced to introduce non-linearity, further improving the model’s fitting ability. During channel interaction, a customized decomposition layer and an element-wise scaling layer are used to redistribute and adjust features across channels, enhancing inter-channel information exchange. Finally, another 1×1 convolution maps the features back to their original dimension. Our approach significantly reduces information redundancy and enhances the model’s capability to represent complex features. 
	
	The extracted features are sent to a global average pooling layer to yield global feature representation, \textcolor{blue}{followed by} a classification layer with a softmax function that ultimately produce the final classification results. Based on this, SAEViT effectively addresses the inefficient feature processing mechanism in traditional ViT, which also provides an efficient and lightweight solution for vision tasks. 
	
	\subsection{Sparsely Aggregated Attention}
	\textcolor{blue}{In vision tasks, adjacent image patches often have high redundancies (i.e. they share high similarities). Therefore, we replace the standard attention block with SAA in our lightweight model to mitigate the redundancy and also to capture long-range spatial dependencies,} so that the model will be able to filter out and integrate tokens that carry similar information(i.e. reduce the number of tokens and computation costs). SAA is featured by spatial sampling strategy: by down-sampling the tokens with a rate sr, so that only the features in the key regions are retained. To be specific, the stride of down-sampling is set to sr to aggregate the adjacent tokens with high similarity. Then, each non-overlapping sr × sr window is reduced to 1 token through average pooling. It not only reduces the complexity of attention operation, but also preserves the main information within the window through local feature smoothing, avoiding the loss of key details caused by random sampling. This is similar to MaxViT[23]’s "Sparse Global Attention" and Biformer[33]’s "Dynamic Sparse Routing" approach, both of them reduce computational complexity through spatial down-sampling. However, unlike them, SAA later up-samples the feature map via deconvolution so as to recover the original spatial dimension. Deconvolution takes a small sized input feature map and outputs a larger feature map, which restores the spatial structural information through a learnable convolution kernel (i.e. enlarging the size by inserting zeros between input elements and performing standard convolution). For given sizes of input feature maps, if the stride is set to s, then the intermediate expended feature map could be created by inserting s-1 number of zeros between each row and column of each spatial unit. Then, a learnable convolution kernel with a size of k×k is used to convolve the feature map, while the output size is controlled by boundary padding.
	
    Compared to traditional methods (e.g. bilinear interpolation), the main advantage of deconvolution is that it parameterizes the up-sampling process and makes it learnable. The convolution kernel is not fixed, instead, it is dynamically optimized through back propagation in model training. As a result, deconvolution learns and reconstructs spatial feature patterns that are highly relevant to specifc tasks in an adaptive way. Meanwhile, deconvolution also inherits the key characteristics of standard convolution operation: Local connectivity and weight sharing. This means that each output pixel is only generated by weighted combinations of elements within the corresponding local neighborhood in the input feature map, and the weights of the same set of convolution kernels slide across the entire input space for reuse. Because of this, deconvolution has a context-aware feature reconstruction ability. This ability not only increases the spatial dimension of feature maps, but also restores important details based on the local context of input data.
	
\begin{figure*}[htbp]
		\centering
		\includegraphics[width=0.85\textwidth]{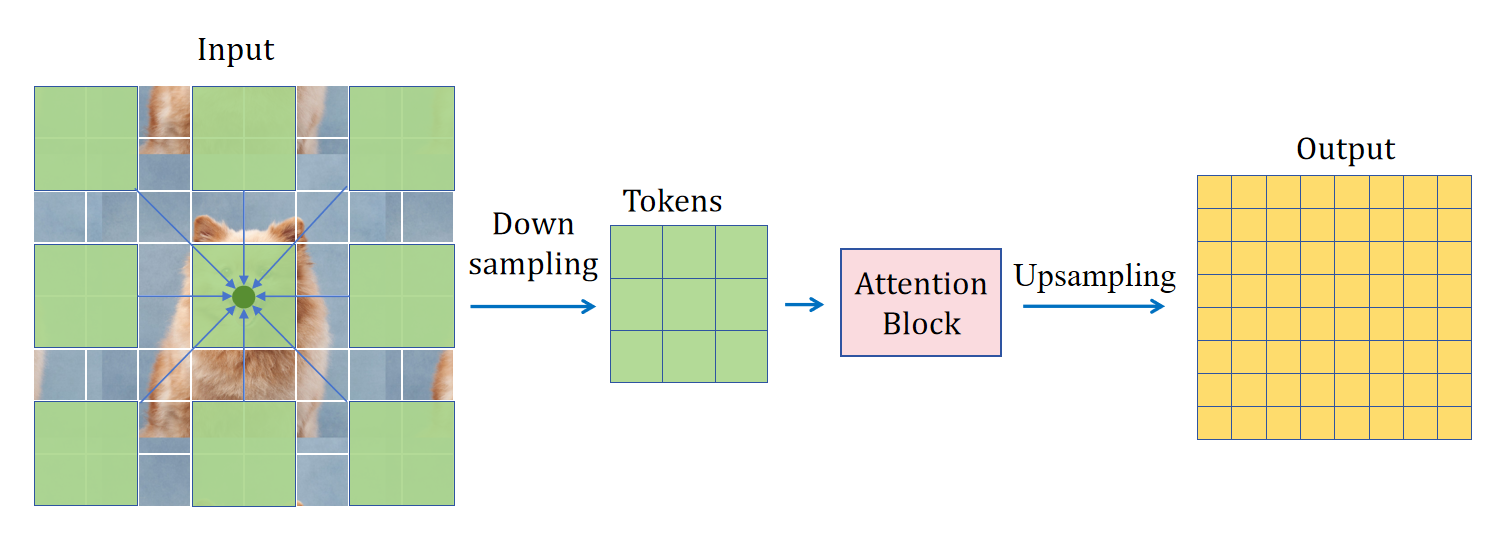}
		\caption{The Feature Scaling Process of SAA. The input is first down-sampled to retain only informative tokens based on redundancy, which is then processed by attention block. Finally, it is up-sampled via deconvolution to reconstruct the full-resolution output.}
		\label{fig:figssa}
		\vspace{10pt}
	\end{figure*}

	SAA is comprised of a reversible sparsity mechanism that is quite different from traditional fixed window or uniform sampling methods. The calculation flow is shown in Fig. \ref{fig:figssa}. 
	
	Assume the input feature map to the attention block at each stage is written as $X_{in} \in R ^{H\times W \times C}$, where H, W, and C represent the height, width, and number of channels, respectively. The spatial down-sampling rate is set to sr (i.e. a token could be aggregated for every $sr \times sr$ windows). Let the aggregation operation be denoted as Aggregation Down-sampling (ADS):
	\begin{equation}
		ADS(X) = AvePooling_{sr \times sr}(X_{in})
	\end{equation}
	\vspace{-20pt}
	\begin{equation}
		X_{1} = ADS(Norm(X_{in}))
	\end{equation}
	Here, $AvePooling$ refers to an average pooling operation determined by the down-sampling rate at each layer of the network backbone, which balances the performances and computation overhead. The down-sampling rate aims to balance the computational complexity and model performance and is controlled by pre-defined model parameters. When $sr>1 $, a down-sampling operation will be performed on the input feature map. Although it reduces computational load, it inevitably causes information loss. We thereby set a reasonable stride of the pooling operation so as to control the down-sampling size. The sparsely down-sampled Q, K, and V are then further processed through the attention module as follows: 
	\begin{equation}
		X_{2} = Attn(X_{1})
	\end{equation}
	Here $Attn$ represents the conventional attention computation. 
	
	The parameter  plays a critical role in both down-sampling and subsequent up-sampling operations. When $sr>1$, after self-attention computation, a deconvolution layer is employed to up-sample the feature map, restoring it to its original spatial dimensions. The deconvolution layer is configured with a kernel size =  $sr$ and stride = $sr$, ensuring that the up-sampled feature map matches the size of the original input feature map. Let $L_{LocalProp}$ denote the deconvolution layer with kernel size and stride equal to sr. In deconvolution operation, the input feature maps are divided into C groups according to the channel dimension. Then the final output $X_{out}$ is generated via:
	\begin{equation}
		X_{out} = L_{LocalProp}(X_{2}) + X_{in} \quad  X \in R^{C \times H \times W}
	\end{equation}
		\vspace{-20pt}
	\begin{equation}
		L_{LocalProp} = Deconv(kernel\_size = sr , stride = sr)
	\end{equation}
	Based on this, the computation cost of attention operation can be effectively reduced to 1/4 of the original, thereby enhancing the computation efficiency. Further analysis will be provided in Section 4.4.2 comparing the number of parameters, the computation costs and CPU/GPU throughput among several classical attention blocks.

	\subsection{Channel Interaction Feedforward Network Layer}
\begin{figure*}[!htb]
		\centering
		\subfloat[FFN]{
			\includegraphics[width=0.37\columnwidth]{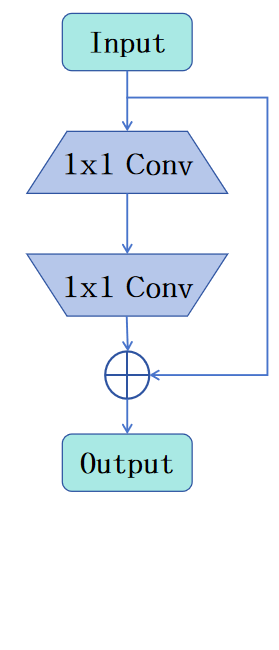}
		}\hspace{1cm}
		\subfloat[DWSFFN]{
			\includegraphics[width=0.37\columnwidth]{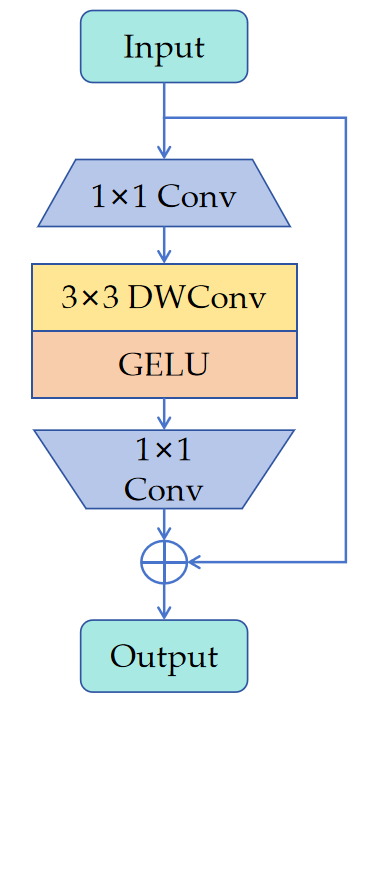}
		}\hspace{1cm}
		\subfloat[CIFFN]{
			\includegraphics[width=0.9\columnwidth]{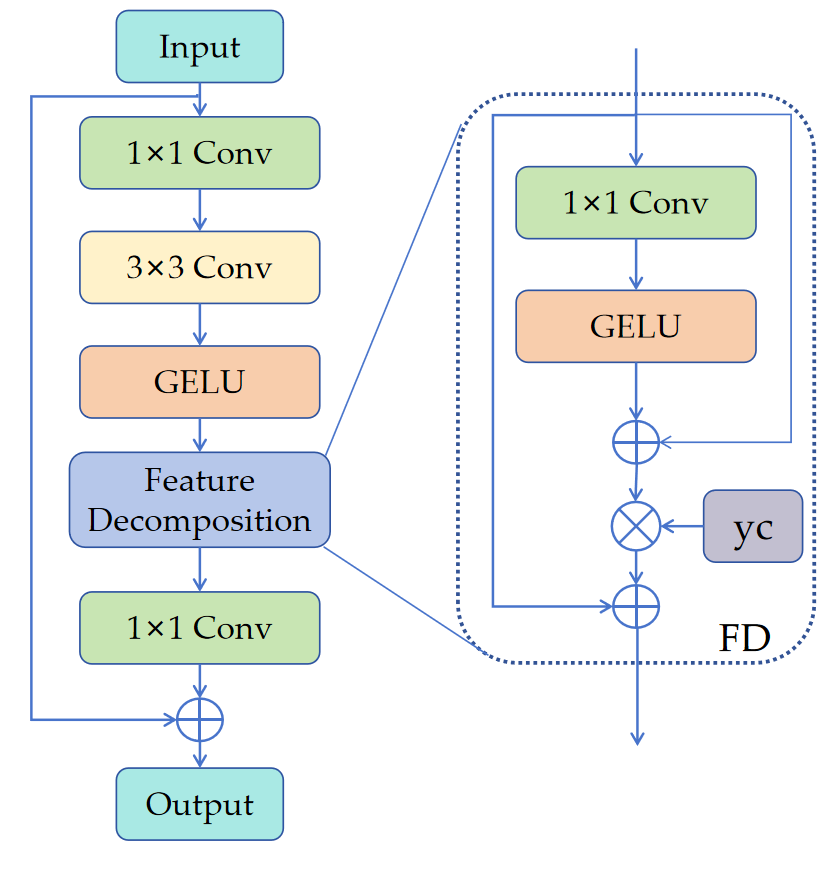}
		}
		\caption{Comparison of structures among FFN, the depth-wise separable FFN, and CIFFN.}
		\vspace{10pt}
		\label{fig:fig2}
	\end{figure*}

	A standard FFN is typically composed of fully connected layers. However, redundancy exists across different channels when dealing with high-dimensional features. In addition, the linear transformation in fully connected layers applies the same operation to all channels, lacking effective screening mechanism and integration of inter-channel information. As a result, the model may focus on some unimportant information during training while ignoring the correlations between channels, \textcolor{blue}{reducing} the model’s expressive power and generalization abilities. To address this issue, most existing methods directly implant channel enhancement modules, such as the Squeeze-and-Excitation (SE) module, into the FFN. Meanwhile, the non-linear transformations in conventional FFN often apply activation functions to the outputs of fully connected layers, which may only capture insufficient multi-scale and hierarchical features under complex feature distributions. Under such circumstances, we propose CIFFN, which consists of convolutional layers for channel mapping, DWConv for local information extraction, and a feature decomposition module for decomposing and recombining features.
	A comparison has been made among FFN, depth-wise separable (DWS) FFN and CIFFN in Fig. \ref{fig:fig2}. The computation flow of CIFFN is expressed as:
	\begin{equation}
		FD(X) = y_{c} \times (X+Conv(X))+X
	\end{equation}
	\vspace{-20pt}
	\begin{equation}
		CIFFN(X) = Conv_{2}(FD(Conv_{1}(X))+X
	\end{equation}
	where FD refers to feature decomposition module.  $y_{c}$ is a learnable channel-wise scaling factor which is initialized to 0. The convolution operation (Conv), is used for feature compression, reducing the number of channels in the input. Given the input feature map , the computation flow of CIFFN is expressed as:
	\begin{equation}
		X_{1} = GELU(DWConv(Conv_{1}(Norm(X_{in}))
	\end{equation}
	\vspace{-20pt}
	\begin{equation}
		X_{2} = FD(X_{1})
	\end{equation}
	\vspace{-20pt}
	\begin{equation}
		X_{out} = Conv2(X_{2}) + X_{in}
	\end{equation}
	It employs a 3×3 DWConv to process the features. Subsequently, a 1×1 convolution layer, referred to as $Conv_{1}$, is utilized to compress the input channels, effectively reducing the dimensionality of the feature map. Following this, another 1×1 convolution layer ($Conv_{2}$) is applied to restore the channel dimensions to their original size. 
	
	Our CIFFN effectively addresses the incompatibility problem by unifying local perception and context aggregation capabilities. It leverages parallel DWSConv to capture interactions across different FFN layers, which captures more interaction information from the intermediate-layers and generates discriminative multi-order representations with minimum computational cost. Furthermore, it mitigates inter-channel information redundancy, enhancing channel efficiency and the model's representational capacity. It also redistributes the channel features in an adaptive way through the lightweight design. Compared to standard FFNs and DWS FFNs, CIFFN achieves superior performance with lower computation cost. 
	
	\subsection{The configuration of the proposed SAE Block}
	In this section, we will discuss the structural details of SAEViT-XS and SAEViT-T. For the i-th stage, assume the output channel dimension is C, "head" denotes the number of heads, "n" refers to the number of blocks. Suppose "sr" is the down-sampling rate for SAA. Then the model structure could be expressed by Table \ref{tab:tabartch}. We select sr based on the characteristics of the feature maps retained in each layer. The parameters should be suitable for the size of the network and feature maps, which also balance information preservation and computational overhead. 
	\begin{table*}[H]
		\caption{The structure of the proposed SAE Block.}
		\centering
		\begin{tabular}{ccccc}
			\toprule
			&Output size & Layer Name& SAEViT-T &SAEViT-XS\\
			\midrule
			Stage1&	112*112&	Conv-stem&	C=24&	C=24\\
			Stage2&	56*56&	SAEBlock&	C=48, head=2,sr=8,n=1&	C=48,head=2,sr=8,n=1\\
			Stage3&	28*28&	SAEBlock&	C=96,head=4,sr=4,n=1&	C=96,head=4,sr=4,n=2\\
			Stage4&	14*14&	SAEBlock&	C=192,head=8,sr=2,n=3&	C=192,head=8,sr=2,n=5\\
			Stage5&	7*7	&SAEBlock&	C=384,head=16,sr=1,n=2	&C=384,head=16,sr=1,n=3\\
			\bottomrule
		\end{tabular}
		\label{tab:tabartch}
	\end{table*}
	In the shallow stage (with 56 × 56 resolution), a larger sampling rate (such as sr=8) is used to actively reduce redundant tokens, while in the deep stage (such as 7 × 7), sr=1 is used to preserve detailed features. As described in Section 3.1, these stride sizes are preset to achieve a reasonable balance between accuracy and FLOPs (within 1.5G). 
	
	The Conv-stem is responsible for data preprocessing. To construct a Conv-stem, we use a combination of convolution operations. Firstly, we use a 3x3 convolution for down-sampling, which not only reduces the spatial dimension of the data, but also extracts basic features from the image. Next, we use 2 convolution blocks with kernel size of 3, stride of 1, and padding of 1. These 2 convolution blocks further refine and enrich features through layer-wise convolution, so that the processed data has high-quality feature expression, which is then sent to the next SAE.

	\section{Experiments}
	In this section, multiple visual recognition experiments have been conducted to verify the effectiveness of our proposed model. First, our model is pre-trained on ImageNet-1K\cite{deng2009imagenet} for image classification task, which is compared with other popular in terms of accuracy and computational cost. Besides, the performances for object detection and segmentation are tested on COCO \cite{lin2014microsoft} and ADE20K\cite{zhou2017scene} respectively. The processing speeds for feature maps are analyzed and compared with other competitors. 
	
	\subsection{Image Classification}
	ImageNet-1K contains approximately 1.28M training images and 50K validation images across 1K categories, which cover a wide variety of objects and scenes, providing sufficient and diverse data resources for model training and evaluation. Following DeiT\cite{touvron2021training}, we utilize AdamW optimizer for fine-tuning, and set the batch size to 256, weight decay coefficient to $5 \times 10^{-2}$ and set momentum to 0.9. The model is trained 300 epochs with an initial learning rate of $1 \times 10^{-3}$, and a linear warm up strategy is applied during the first 5 epochs. After warm up, the learning rate will be adjusted following a cosine decay schedule. Data augmentation techniques are adopted, including random cropping, random horizontal flipping, random erasing, and label smoothing. During training, images are randomly cropped to 224×224 pixels. All experiments are conducted on 2×NVIDIA 3090 GPUs. 
	
	\textbf{Results:} A comparison of classification results are made between our model and other popular methods, including MobileNetV2\cite{sandler2018mobilenetv2} and EfficientNet\cite{tan2019efficientnet} as well as PVT\cite{wang2021pyramid}, PVTv2\cite{wang2022pvt}, and DeiT\cite{touvron2021training}. All methods are roughly divided into 2 groups based on their number of parameters and floating-point operations (FLOPs). As shown in Table \ref{tab:tab1}, under comparable network size, \textcolor{red}{SAEViT-T and SAEViT-XS achieve Top-1 accuracies of 76.3\% and 79.6\% on ImageNet-1K with only 0.8 GFLOPs and 1.3 GFLOPs, respectively., which outperform other lightweight models such as MobileNetV2\cite{sandler2018mobilenetv2}, EfficientNet\cite{tan2019efficientnet}, PVT\cite{wang2021pyramid}, and DeiT-T\cite{touvron2021training}. Notably, SAEViT-XS surpasses EfficientNet-B1 (79.1\% with 0.7 GFLOPs) by 0.5\% in accuracy while using marginally more computation, demonstrating its superior efficiency-accuracy trade-off. The performance gain can be attributed to the effective integration of convolutional inductive biases and sparse self-attention, which enhances local feature extraction and reduces redundancy without sacrificing global contextual understanding. Additionally, SAEViT-T realizes 0.5\% improvement in accuracy over SBCFormer-XS\cite{lu2024sbcformer}, while SAEViT-XS leads SBCFormer-X\cite{lu2024sbcformer} by 1.9\% in accuracy. Similarly, SAEViT achieves higher precision than Mobile-Former. The reason is that SBCFormer adopts more aggressive spatial downsampling operation to pursue lower CPU latency, while Mobile-Former\cite{chen2021mobileformer} employs a parallel MobileNet+Transformer architecture with very few tokens to prioritize efficiency. In contrast, our model strikes balanced local-global design under lightweight framework, which makes stronger representational capabilities and Top-1 accuracy improvement.} 
	
	\begin{table}[tb]
		\caption{Comparison of Classification results on ImageNet-1K.}
		\centering
		\begin{tabularx}{\linewidth}{>{\raggedright\arraybackslash}c c c c}
			\toprule
			Model & \makecell{Params\\(M)} & GFLOPs & \makecell{Accuracy\\(\%)} \\
			\midrule
			PVTv2-b0~\cite{wang2022pvt} & 3.4 & 0.6 & 70.5 \\
			MobileNetV2×1.0~\cite{sandler2018mobilenetv2} & 3.5 & 0.3 & 71.8 \\
			UniFormer-Tiny~\cite{li2022uniformer} & 3.9 & 0.6 & 74.1 \\
			MSCViT-T~\cite{zhang2025mscvit} & 3.8 & 0.5 & 73.9 \\
			DeiT-T~\cite{touvron2021training} & 5.7 & 1.3 & 72.2 \\
			MobileNetV3 0.75~\cite{howard2019searching} & 4.0 & 0.2 & 73.3 \\
			EdgeViT-XXS~\cite{pan2022edgevits} & 4.1 & 0.6 & 74.4 \\
			CPVT-Ti-GAP~\cite{chu2021conditional} & 6.0 & 1.3 & 74.9 \\
			SBCFormer-XS~\cite{lu2024sbcformer} & 5.6 & 0.7 & 75.8 \\
			SDGFormer-T~\cite{wen2023sdgformer} & 6.3 & 0.8 & 76.2 \\
			Mobile-Former-151M~\cite{chen2021mobileformer} & 7.6 & 0.2 & 75.2 \\
			CaiT-xxs12-withDW\cite{ZHANG2025} & 6.6 & 1.3 & 75.8 \\
			Incepformer-S\cite{MENG2025} & 16.07 & 9.41 & 75.3 \\
			\textbf{SAEViT-T} & 6.0 & 0.8 & \textbf{76.3} \\
			\midrule
			MobileNetV2×1.4~\cite{sandler2018mobilenetv2} & 6.1 & 0.6 & 74.7 \\
			EdgeViT-XS~\cite{pan2022edgevits} & 6.7 & 1.1 & 77.5 \\
			MSCViT-XS~\cite{zhang2025mscvit} & 7.7 & 1.0 & 77.3 \\
			PVT-T~\cite{wang2021pyramid} & 13.2 & 1.9 & 75.1 \\
			SBCFormer-X~\cite{lu2024sbcformer} & 8.5 & 0.9 & 77.7 \\
			SDGFormer-S~\cite{wen2023sdgformer} & 9.1 & 1.6 & 77.6 \\
			EfficientNet-B1~\cite{tan2019efficientnet} & 7.8 & 0.7 & 79.1 \\
			Mobile-Former-294M~\cite{chen2021mobileformer} & 11.4 & 0.3 & 77.9 \\
			Mobile-Former-508M~\cite{chen2021mobileformer} & 14.0 & 0.5 & 79.3 \\
			CaiT-xxs24-withDW\cite{ZHANG2025} & 12.0 & 2.5 & 79.6 \\
			Incepformer-M\cite{MENG2025} & 30.96 & 19.2 & 77.5 \\
			\textbf{SAEViT-XS} & 8.9 & 1.3 & \textbf{79.6} \\
			\bottomrule
		\end{tabularx}
		\label{tab:tab1}
	\end{table}

	\subsection{Object Detection and Instance Segmentation}
	Object detection has been conducted on COCO dataset, where the training set contains 118k images and the validation set contains 5k images. We utilize 2 commonly used detectors (RetinaNet\cite{lin2017focal} and Mask R-CNN\cite{he2017mask}) to validate the effectiveness of SAEViT as a backbone network. 
	
	Before training, the model initializes the SAEViT backbone with pre-trained weights from the ImageNet dataset and applies Xavier initialization to the newly added layers. Our model is trained on 2 RTX 3090 GPUs, with a batch size of 8 and AdamW optimizer, where the initial learning rate is set to $1 \times 10^{-4}$. Following common practices, the model is fine-tuned with 1× schedule (i.e. 12 epochs). The shorter side of input images is resized to 800 pixels, while the longer side does not exceed 1333 pixels. For testing, the shorter side of input images is fixed at 800 pixels. 
	
	For RetinaNet, $AP_{50}$ refers to the mean Average Precision (mAP) with an Intersection over Union (IoU) threshold of 0.5, and $AP_{75}$ corresponds to mAP with an IoU threshold of 0.75. $AP_{S}$, $AP_{M}$, $AP_{L}$ denote AP values for small, medium, and large objects, respectively. For Mask R-CNN, box mAP and mask mAP are used as the primary evaluation metrics. 
	
	\textbf{Results:} The experimental results on the COCO dataset are detailed in Tables \ref{tab:tab2} and \ref{tab:tab3} by using RetinaNet and Mask R-CNN respectively. Apparently, SAEViT ranks first in each group. Especially, SAEViT-T and SAEViT-XS achieve 39.5 AP and 41.8 AP respectively, surpassing other competitors with safe margins. Similarly, using Mask R-CNN detector, SAEViT-T and SAEViT-XS achieve 40.1 AP and 42.6 AP respectively, which exhibits strong object localization and segmentation ability under complex scenes. \textcolor{red}{PoolFormer-S12\cite{yu2022metaformer}, PVTv2-B0\cite{wang2022pvt} and EdgeViT-XS\cite{pan2022edgevits} only achieve 36.2 AP, 37.2 AP and 38.7AP respectively. EdgeViT-XS adopts a local-global-local delegate-token mechanism, which reduces the cost of self-attention while performing strong aggregation and sparsification operations on spatial information. In contrast, our SAA uses learnable upsampling (deconvolution + skip connections) to recover more high-frequency details, which makes it more advantageous in detecting small object. On the other hand, CIFFN enhances inter-channel interaction, allowing the model to better discriminate object features from cluttered backgrounds. The complementary function of SAA+CIFFN forms a strong backbone, providing richer and more discriminative multi-scale feature representations. 
	Similarly, SAEViT-XS achieves 41.8 AP with RetinaNet, surpassing SBCFormer-L\cite{lu2024sbcformer}(41.1 AP) and EdgeViT-XS\cite{pan2022edgevits}(40.6 AP). The improvement is more prominent by using Mask R-CNN, where SAEViT-XS reaches 42.6 box AP and 43.5 mask AP, outperforming RepViT-M1.1\cite{wang2024repvit}(39.8 AP) and EfficientFormer-L3\cite{li2022efficientformer}(41.4 AP). These results confirm the superiority of our multi-scale hierarchical design in enhancing feature representation for both object localization and segmentation. The ability to preserve spatial details through SAA and enrich channel interactions via CIFFN allows our model to excel in complex scenes with overlapping objects and varied scales.} 
	
	\begin{table}[tb]
		\setlength{\tabcolsep}{3pt}
		\caption{Performance of SAEViT on RetinaNet.}
		\centering
		\begin{tabular*}{\linewidth}{@{\extracolsep\fill}ccccccc@{\extracolsep\fill}}
			\toprule
			Model     &AP&$AP_{50}$&$AP_{75}$&$AP_{S}$&$AP_{M}$&$AP_{L}$\\
			\midrule
			PoolFormer-S12\cite{yu2022metaformer}&36.2&56.2&38.2&20.8&39.1&48.0\\
			PVTv2-B0\cite{wang2022pvt}&37.2&57.2&39.5&23.1&40.4&49.7\\
			EdgeViT-XXS\cite{pan2022edgevits}&38.7&59.0&41.0&22.4&42.0&51.6\\
			SBCFormer-B\cite{lu2024sbcformer}&39.3&59.8&41.5&21.9&\textbf{42.7}&\textbf{53.3}\\
			\textbf{SAEViT-T}&\textbf{39.5}&\textbf{60.2}&\textbf{41.8}&\textbf{23.8}&42.5&53.1\\
			\midrule
			PoolFormer-S24\cite{yu2022metaformer}&38.9&59.7&41.3&23.3&42.1&51.8\\
			EdgeViT-XS\cite{pan2022edgevits}&40.6&61.3&43.3&25.2&43.9&54.6\\
			SBCFormer-L\cite{lu2024sbcformer}&41.1&62.3&43.3&24.7&44.3&\textbf{56.0}\\
			\textbf{SAEViT-XS}&\textbf{41.8}&\textbf{62.5}&\textbf{44.5}&\textbf{25.6}&\textbf{44.9}&55.8\\
			\bottomrule
		\end{tabular*}
		\label{tab:tab2}
	\end{table}
	
\begin{figure*}[!htb]
		\centering
		\includegraphics[width=0.9\linewidth]{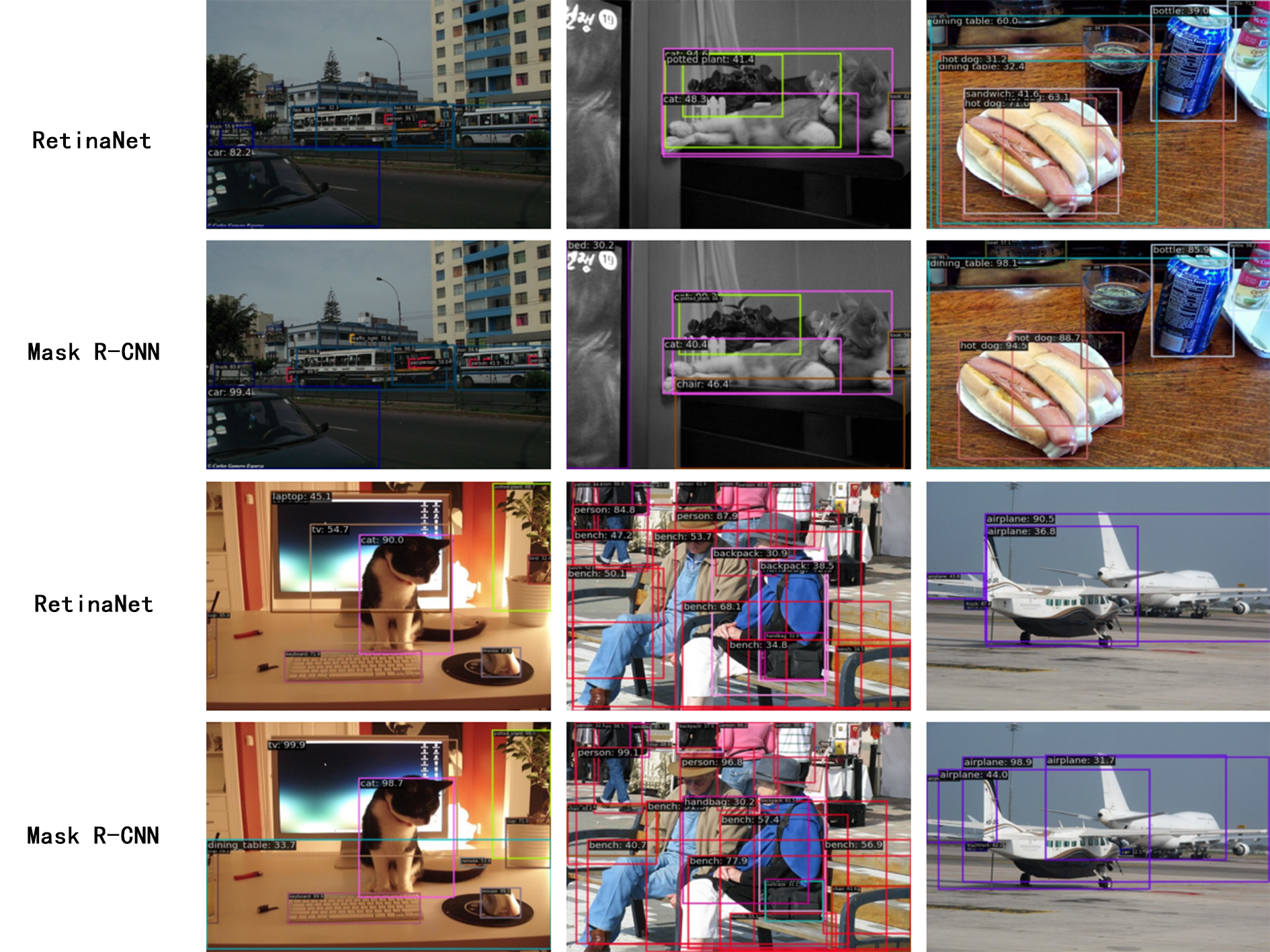}
		\caption{The detection results on COCO dataset.}
		\label{fig:fig3}
		\vspace{10pt}
	\end{figure*}

	%%%%%%%%%%%%%%%%%%%%%%%%%%%%%%%%%%%%%%%%%%%%
	\begin{table}[tb]
		\setlength{\tabcolsep}{3pt}
		\caption{Performance of SAEViT on Mask R-CNN.}
		\centering
		\begin{tabular*}{\linewidth}{@{\extracolsep\fill}ccccccc@{\extracolsep\fill}} 
			\toprule
			Model     &$AP_{b}$&$AP^{b}_{50}$&$AP^{b}_{75}$&$AP^{m}$&$AP^{m}_{50}$&$AP^{m}_{75}$\\
			\midrule
			EfficientFormer-L1\cite{li2022efficientformer}&37.9&60.3&41.0&35.4&57.3&37.3\\
			\textbf{SAEViT-T}&\textbf{40.1}&\textbf{61.8}&\textbf{43.2}&\textbf{37.6}&\textbf{59.5}&\textbf{40.1}\\
			\midrule
			RepViT-M1.1\cite{wang2024repvit}&39.8&61.9&43.5&37.2&58.8&40.1\\
			EfficientFormer-L3\cite{li2022efficientformer}&41.4&63.9&44.7&38.1&61.0&40.4\\
			\textbf{SAEViT-XS}&\textbf{42.6}&\textbf{64.3}&\textbf{46.2}&\textbf{40.3}&\textbf{62.8}&\textbf{43.5}\\
			\bottomrule
		\end{tabular*}
		\label{tab:tab3}
	\end{table}
	%%%%%%%%%%%%%%%%%%%%%%%%%%%%
	\textbf{Visualizations:} Fig. \ref{fig:fig3} shows detection results by using different detectors. The $1^{st}$ and the $3^{rd}$ columns shows the results of SAEViT-XS by using RetinaNet, the $2^{nd}$ and the $4^{th}$ columns show the results by using Mask R-CNN. Generally, \textcolor{red}{SAEViT-based detectors maintain high precision and recall, which have successfully identified most} commonly known objects(except that RetinaNet has some missed detections, while Mask R-CNN made some duplicated detections). \textcolor{red}{Since our model preserves the fine-grained details through sparse attention and enhances channel interactions, which greatly improve the detection results.} 
	
	\subsection{Semantic Segmentation on ADE20K}
	The semantic segmentation experiment is conducted on the ADE20K scene parsing dataset. The overall performance is evaluated using Semantic FPN detector. During training, the network is initialized with ImageNet pre-trained weights, and any newly added layers are initialized via Xavier initialization. The AdamW optimizer is used with an initial learning rate of $1 \times 10^{-4}$. Following common practice, our model is trained 80K epochs on a single NVidia 3090 GPU with a batch size of 16. The learning rate is adjusted using a polynomial decay schedule. During training, input images are randomly resized and cropped to 512×512. During testing, the image’s shorter side is resized to 512 pixels. The evaluation metric is mIoU (i.e. mean Intersection over Union). 
	
	\begin{table}[tb]
		\caption{Segmentation Results on ADE20K.}
		\centering
		\begin{tabular*}{\linewidth}{@{\extracolsep\fill}ccc@{\extracolsep\fill}} 
			\toprule
			Model     &Params(M)&mIoU\\
			\midrule
			
			PVTv2-b0\cite{wang2022pvt}&7.6&37.2\\
			PoolFormer-S12\cite{yu2022metaformer}&7.6&37.2\\
			EdgeViT-XXS\cite{pan2022edgevits}&7.9&39.7\\
			\textbf{SAEViT-T}&10.1&\textbf{40.6}\\
			\midrule
			RepViT-m1.1\cite{wang2024repvit}&12.0&40.6\\
			EdgeViT-XS\cite{pan2022edgevits}&10.6&41.4\\
			PVT-S\cite{wang2021pyramid}&28.2&39.8\\
			PoolFormer-S12\cite{yu2022metaformer}&23.2&40.3\\
			EfficientFormer-L1\cite{li2022efficientformer}&12.3&38.9\\
			\textbf{SAEViT-XS}&13.1&\textbf{42.1}\\
			\bottomrule
		\end{tabular*}
		\label{tab:tab4}
	\end{table}
	
	\textbf{Results:} As shown in Table \ref{tab:tab4}, SAEViT achieves appealing performance on ADE20K semantic segmentation benchmarks. \textcolor{red}{SAEViT-T achieves 40.6 mIoU with only 10.1M parameters, while SAEViT-XS attains 42.1 mIoU with 13.1M parameters, outperforming PVTv2\cite{wang2022pvt}, PoolFormer\cite{yu2022metaformer}, and EdgeViT \cite{pan2022edgevits} under the same level of parameter constraints. PVTv2 emphasizes hierarchical token processing and PoolFormer focuses on a minimalist token-mixer, which all ignore the recovery and fusion of multi-scale details with reduced dimensionality. In contrast, SAEViT has strong capability in capturing and reconstructing semantic contexts across diverse scenes. The performance improvement can be largely attributed to the model’s multi-scale feature aggregation mechanism enabled by SAA and enhanced inter-channel communication through CIFFN. Considering that precise pixel-level classification and context integration are critical for segmentation tasks, SAA preserves structural details even after downsampling via learnable deconvolution, maintaining boundary information. Meanwhile, CIFFN promotes more discriminative feature learning through channel-wise re-assginment, effectively reducing intra-class ambiguity and ensuring segmentation consistency, especially for categories with high visual similarity. Furthermore, the hierarchical pyramid design of SAEViT allows the model to capture features at multiple scales, making it particularly suited for segmenting objects of varying sizes within complex scenes.}
	
	\subsection{GPU Throughput and Latency tests}\label{throughput}
	%In this section, throughput tests are implemented multiple times to measure the model’s average throughput on a given device (i.e. the number of images processed per second). The experimental setup includes a 5s warm up period, random initialization of 224×224 feature maps, and the batch size is set to 256. The tests are conducted on the same AMD Ryzen 7 5800H CPU and NVIDIA RTX 3060 Laptop GPU, where the comparison is made among PVT-T, ShuntedViT-T, CMT-Ti  and our model SAEViT-XS. The results have been shown in Table \ref{tab:tab5}. Our model has the minimum number of parameters and relatively low computation cost, it also achieves the second to the best accuracy and has the highest throughput, which exceeds other methods with large margin. 
	As the key indicator for measuring the real-time inference performance of deep learning models, while device latency directly reflects the response speed of the model in practical application scenarios such as real-time image recognition and autonomous driving decision systems. In this section, throughput and latency tests are implemented multiple times to measure the model’s average throughput (i.e. the number of images processed per second) and single-image inference latency (milliseconds) on a given device. The experimental setup includes a 5s warm-up period, random initialization of 224×224 feature maps, and batch sizes \textcolor{blue}{are} set to 256 for throughput measurement and 1 for latency measurement. \textcolor{blue}{The forward propagation is performed 50 times to ensure fair comparison.} All tests are conducted on the same AMD Ryzen 7 5800H CPU and NVIDIA RTX 3060 Laptop GPU, comparing PVT-T, ShuntedViT-T, CMT-Ti and our model SAEViT-XS. 
	
	\textcolor{red}{Table \ref{tab:tab5} compares the throughput and latency of SAEViT-XS against other lightweight vision Transformers. Among them, SAEViT-XS achieves the highest throughput and the lowest latency on an NVIDIA RTX 3060 GPU with fewer parameters. This efficiency stems from our sparsely aggregated attention mechanism, which reduces redundant computations, and the optimized CIFFN module minimizes inter-channel redundancy without extensive linear projections. As a result, SAEViT has high suitability for real-time applications on edge devices, where both accuracy and inference speed are critical.}
	
	\begin{table*}[tb]
		\setlength{\tabcolsep}{3pt}
		\caption{The comparison of throughputs and device Latency among several popular methods.}
		\centering
		\begin{tabular*}{\linewidth}{@{\extracolsep\fill}cccccc@{\extracolsep\fill}} 
			\toprule
			Model	&Params(M)& GFLOPs	&Throughput& Accuracy(\%)& Latency(ms)\\
			\midrule
			PVT-T\cite{wang2021pyramid}	&13.2	&1.9	&420.55&	75.1&	44.62\\
			ShuntedViT-T \cite{ren2022shunted}&	11.5&	2.1&	168.40&	\textbf{79.8}&	52.57\\
			CMT-Ti\cite{guo2022cmt}&	9.5&	\textbf{0.6}&	496.36	&79.1&60.88\\
			\textbf{SAEViT-XS}&\textbf{8.9}	&1.3&	\textbf{575.89}&	79.6&	\textbf{36.80}\\
			\bottomrule
		\end{tabular*}
		\label{tab:tab5}
	\end{table*}
	
	\subsection{Ablation Studies}
	In this section, we will investigate the effectiveness of the core components of our model, as well as the advantages of SAA and CIFFN.  
	
	\subsubsection{Analysis of the core components in our model}
	In this section, we investigate the impacts of the core components, including SAA, CIFFN and LFE. Table \ref{tab:tab7} shows the classification accuracies by using and without using the 3 components. Specifically, SRA and FFN substitute SAA and CIFFN respectively. As mentioned earlier, LFE is an embedding layer with DWSConv, which achieves smooth noise suppression and refines the input features to the Transformer block. It introduces only 0.1 M parameters and 0.01G computation, but brings 0.6\% Top-1 accuracy improvement. Compared with SRA, SAA achieves 0.5\% higher accuracy, the reason is that the deconvolution operation in SAA recovers more feature information than the spatial reduction in SRA. The common goal of SAA and SRA is to reduce the computational cost of self-attention through spatial downsampling. SRA only performs pooling or linear mapping on key and value tensors. By contrast, SAA processes the entire feature map by performing pooling on the normalized input and fusing it with the original features through skip connections. After attention operation, SAA uses deconvolution for upsampling to restore spatial resolution, which is different from SRA. After sr×sr spatial down-sampling, it reconstructs the spatial structure through deconvolution to preserve high-frequency edges and textures. Although it introduces 0.3M parameters and 0.02G computation, it also improves Top-1 accuracies. CIFFN makes the greatest increase in accuracy than LFE and SAA, as its channel-wise feature selection and integration facilitate more effective feature learning. CIFFN uses channel decomposition and learnable channel-level re-calibration to filter and re-distribute channel-level information. It yields 79.6\% accuracy at the cost of 0.4M parameters and 0.03G computation. It’s worth noting that the 3 modules introduces less than 1M parameters, but improves the accuracy by 2.4\%, which strikes a balance between lightweight design and performance gain. 
	
	\begin{table}[tb]
		\setlength{\tabcolsep}{3pt}
		\caption{Ablation experiment on the core components.}
		\centering
		\begin{tabular*}{\linewidth}{@{\extracolsep\fill}cccc@{\extracolsep\fill}} 
			\toprule
			LFE&	SAA	&CIFFN	&Accuracy(\%) \\
			\midrule
			& & &			77.2\\
			\checkmark&&		&	77.8\\
			\checkmark	&\checkmark	&	&78.3\\
			\checkmark	&\checkmark&	\checkmark&	\textbf{79.6}\\
			\bottomrule
		\end{tabular*}
		\label{tab:tab7}
	\end{table}
	
	\subsubsection{Computational Efficiency of Attention Modules}
	In this section, we compare the comprehensive performances of different attention modules in terms of number of parameters, computation cost, the throughput of both CPU and GPU, so as to investigate the computational efficiency of different attention modules. Specifically, the processing speed of the attention modules is tested on both an AMD Ryzen 7 5800H CPU and an NVIDIA RTX 3060 Laptop GPU. All attention modules adopt identical settings: the input dimension is $X \in X^{H \times W \times C}$, where H=W=56, and C=256. The multi-head attention mechanism has 8 heads, and the input batch size is 128. All other conditions remain identical. As shown in Table \ref{tab:tab6}, SAA has the lowest computation cost and almost the lowest number of parameters, but it realizes the highest CPU and GPU throughputs (i.e. it processes 8187 images/sec on AMD Ryzen 7 5800H and 22942 images/sec on NVidia RTX 3060). This results demonstrate that our SAA has obvious advantage in processing efficiency, especially SAA is more CPU-friendly than other methods.
	
	\begin{table}[tb]
		\setlength{\tabcolsep}{3pt}
		\caption{Comprehensive performance comparison among our proposed SAA and other popular attention modules.}
		\centering
		\begin{tabular*}{\linewidth}{@{\extracolsep\fill}ccccc@{\extracolsep\fill}} 
			\toprule
			Attention Modules	&Params(M)&	FLOPs	& CPU&	 GPU\\
			\midrule
			SRA\cite{wang2021pyramid}	&0.53	&1979.14&	2674&	17345\\
			Cross Attention\cite{dong2022cswin}&	0.27&	1458.72&	5002&	18706\\
			Shifted Window\cite{liu2021swin}&0.26&	900.76&	6058&	13479\\
			HIRO\cite{pan2022fast}&	\textbf{0.25}&	1609.65&	3342&	20772\\
			MSC\cite{zhang2025mscvit}&0.41	&2422.35&	2131&	11123\\
			SAA	&0.26	&\textbf{679.18}&	\textbf{8187}&	\textbf{22942}\\
			\bottomrule
		\end{tabular*}
		\label{tab:tab6}
	\end{table}
	
	\subsubsection{Inter-channel Enhancement Effect of CIFFN}
\begin{figure*}[!htbp]
		\centering
		\subfloat[FFN]{
			\includegraphics[width=0.8\columnwidth]{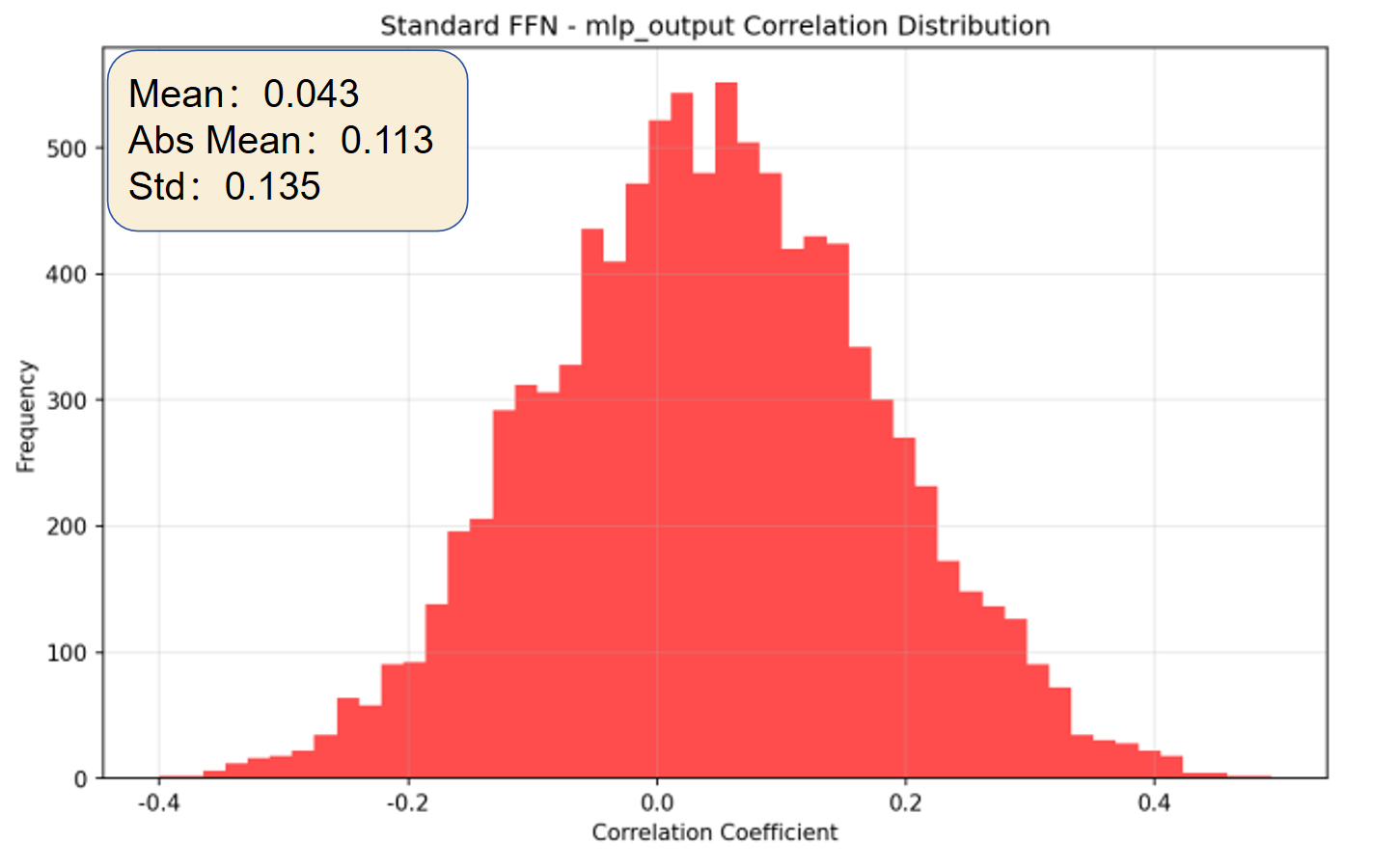}
		}\hspace{1cm}
		\subfloat[DWSFFN]{
			\includegraphics[width=0.8\columnwidth]{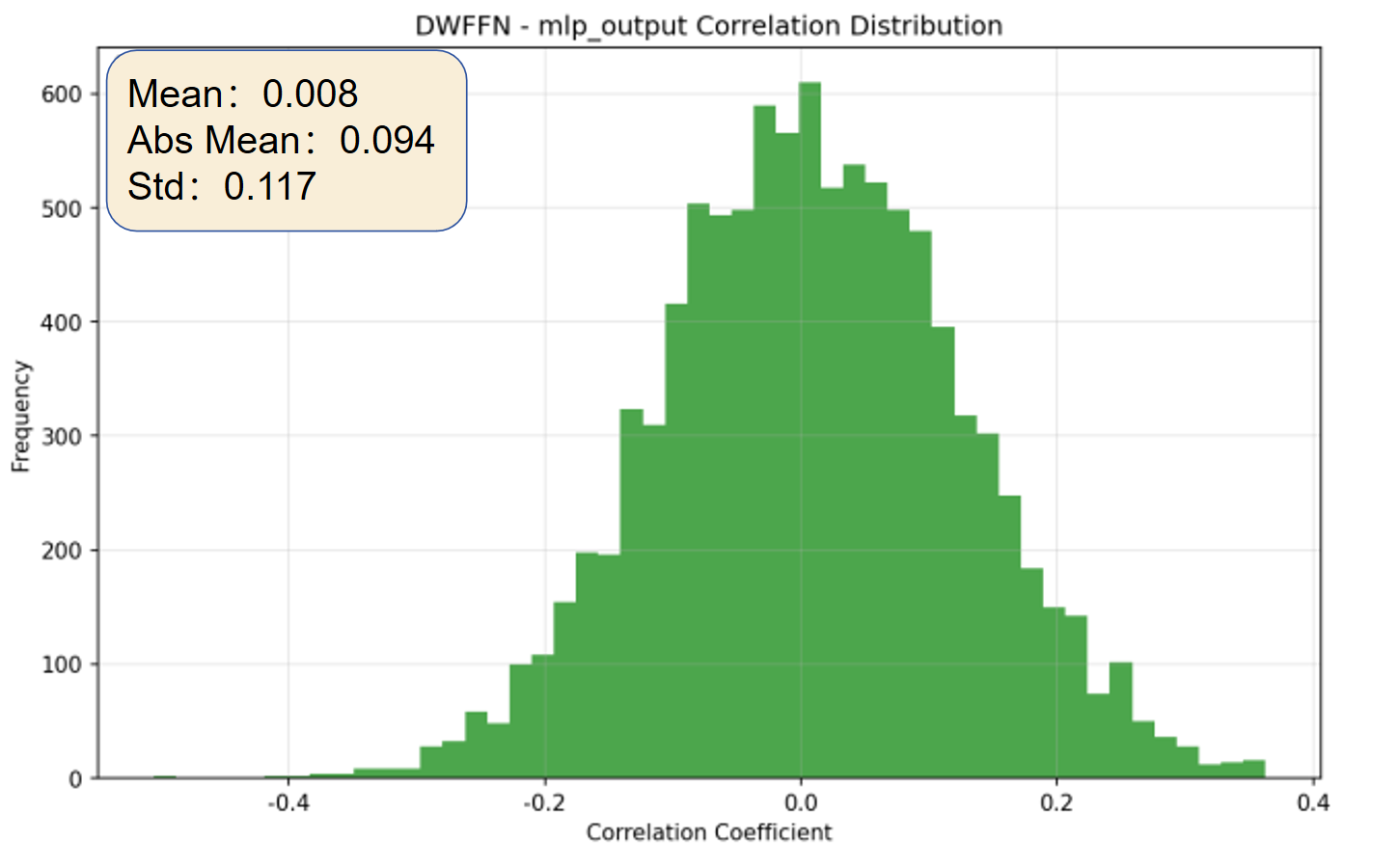}
		}\hspace{1cm}
		\subfloat[CIFFN]{
			\includegraphics[width=0.8\columnwidth]{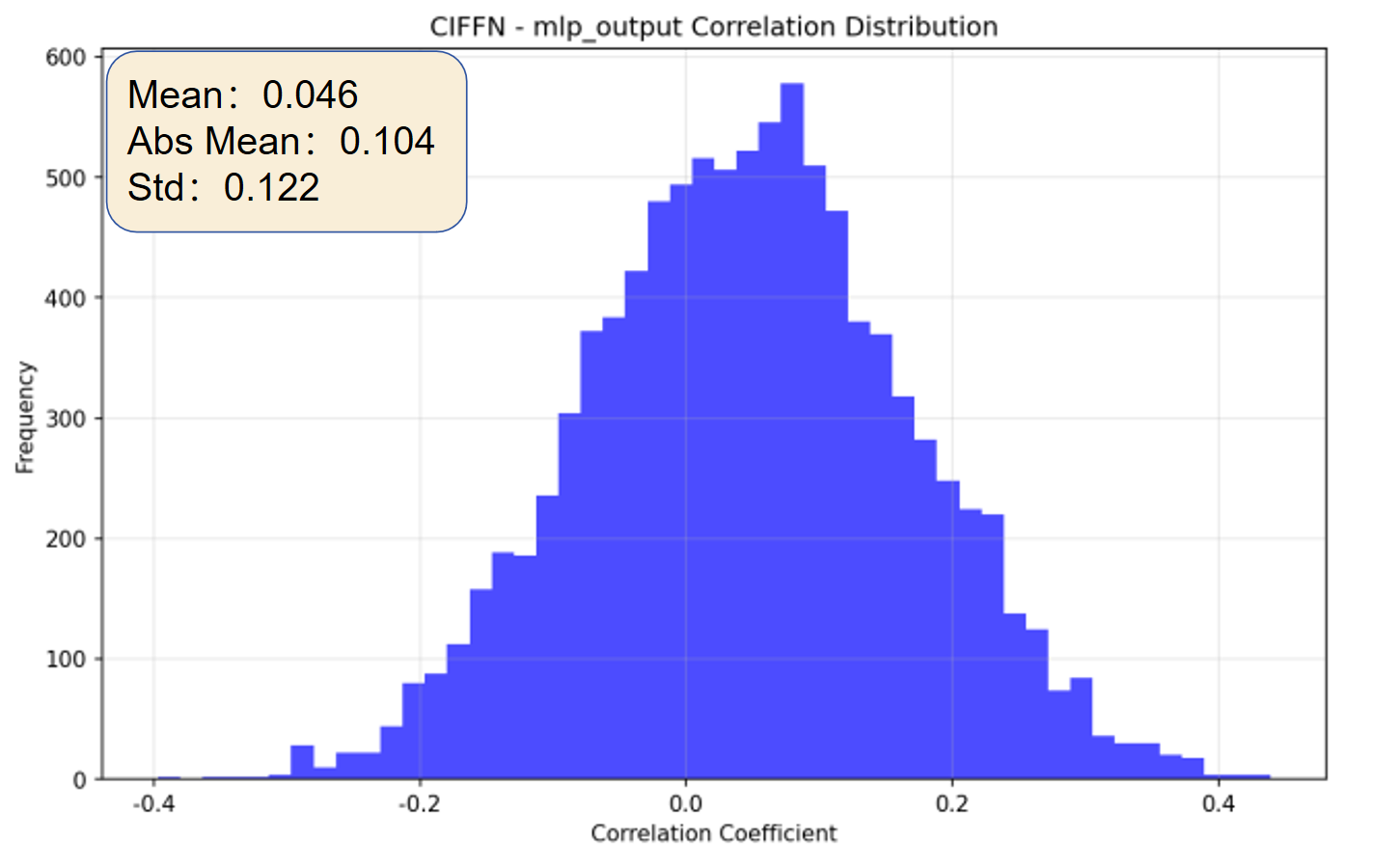}
		}
		\caption{Comparison of correlation coefficient among FFN, DWFFN and CIFFN.}
		\vspace{10pt}
		\label{fig:fig4}
	\end{figure*}

\begin{figure*}[!htbp]
		\centering
		\subfloat[FFN]{
			\includegraphics[width=0.6\columnwidth]{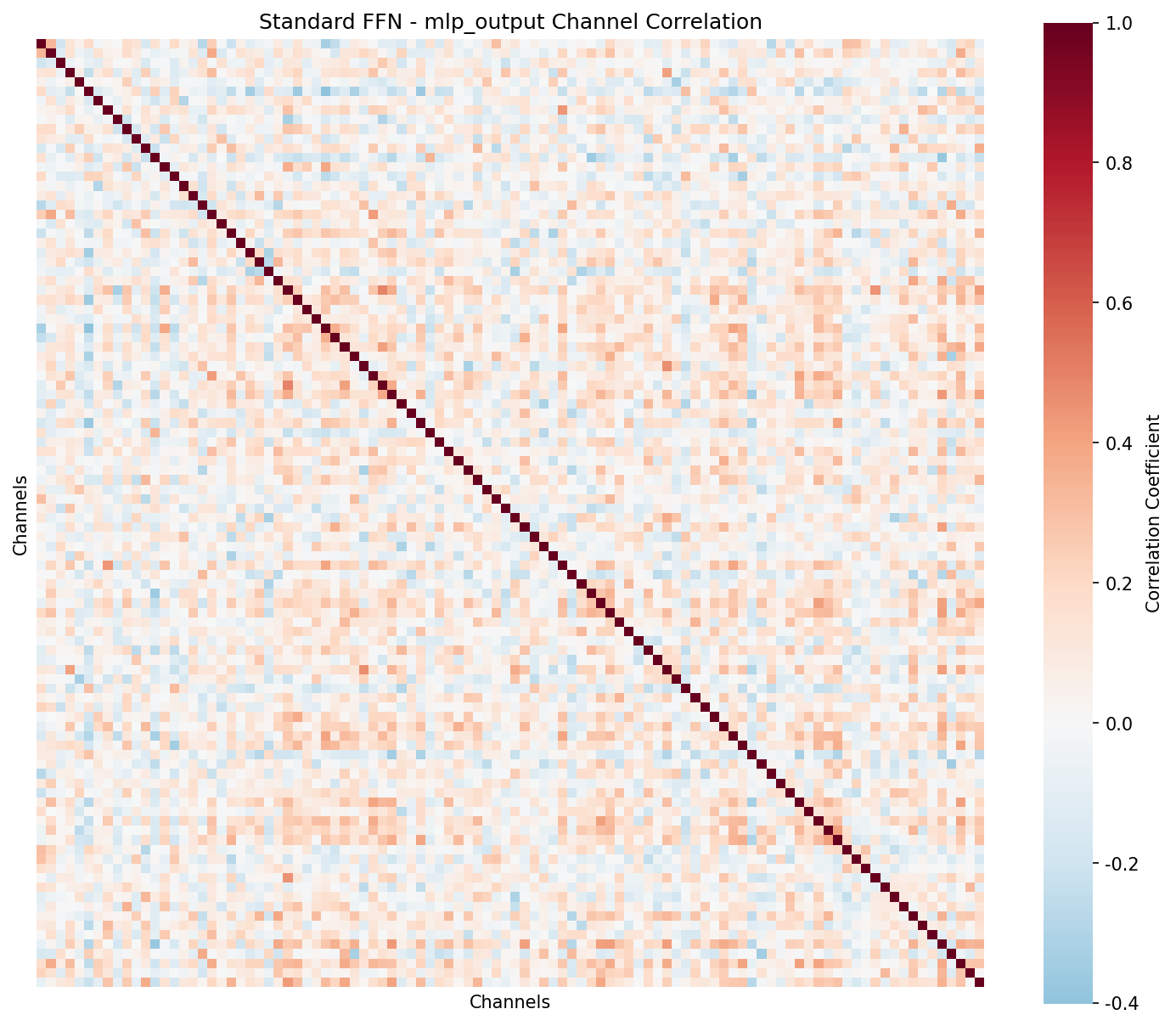}
		}\hspace{0.5cm}
		\subfloat[DWSFFN]{
			\includegraphics[width=0.6\columnwidth]{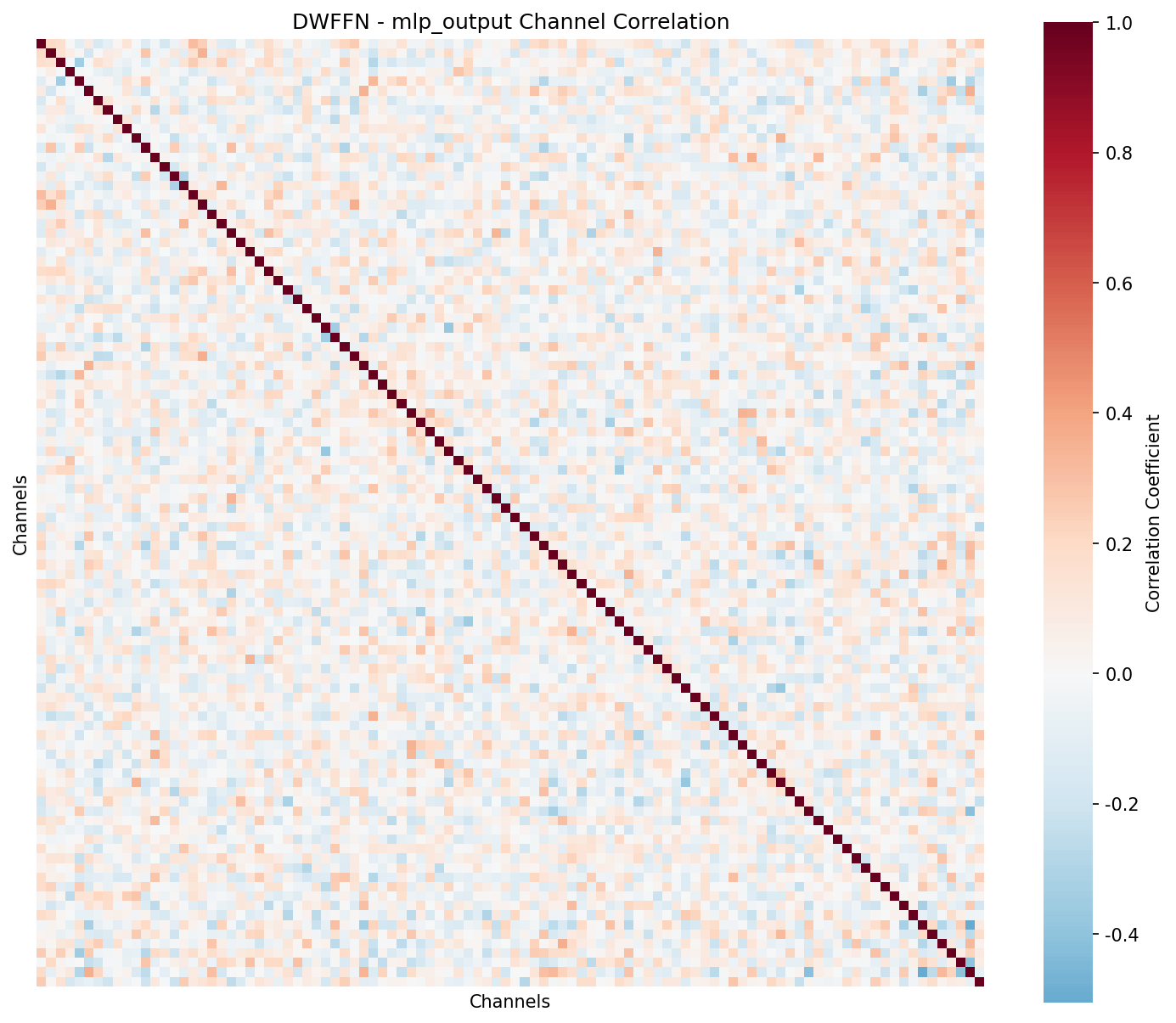}
		}\hspace{0.5cm}
		\subfloat[CIFFN]{
			\includegraphics[width=0.6\columnwidth]{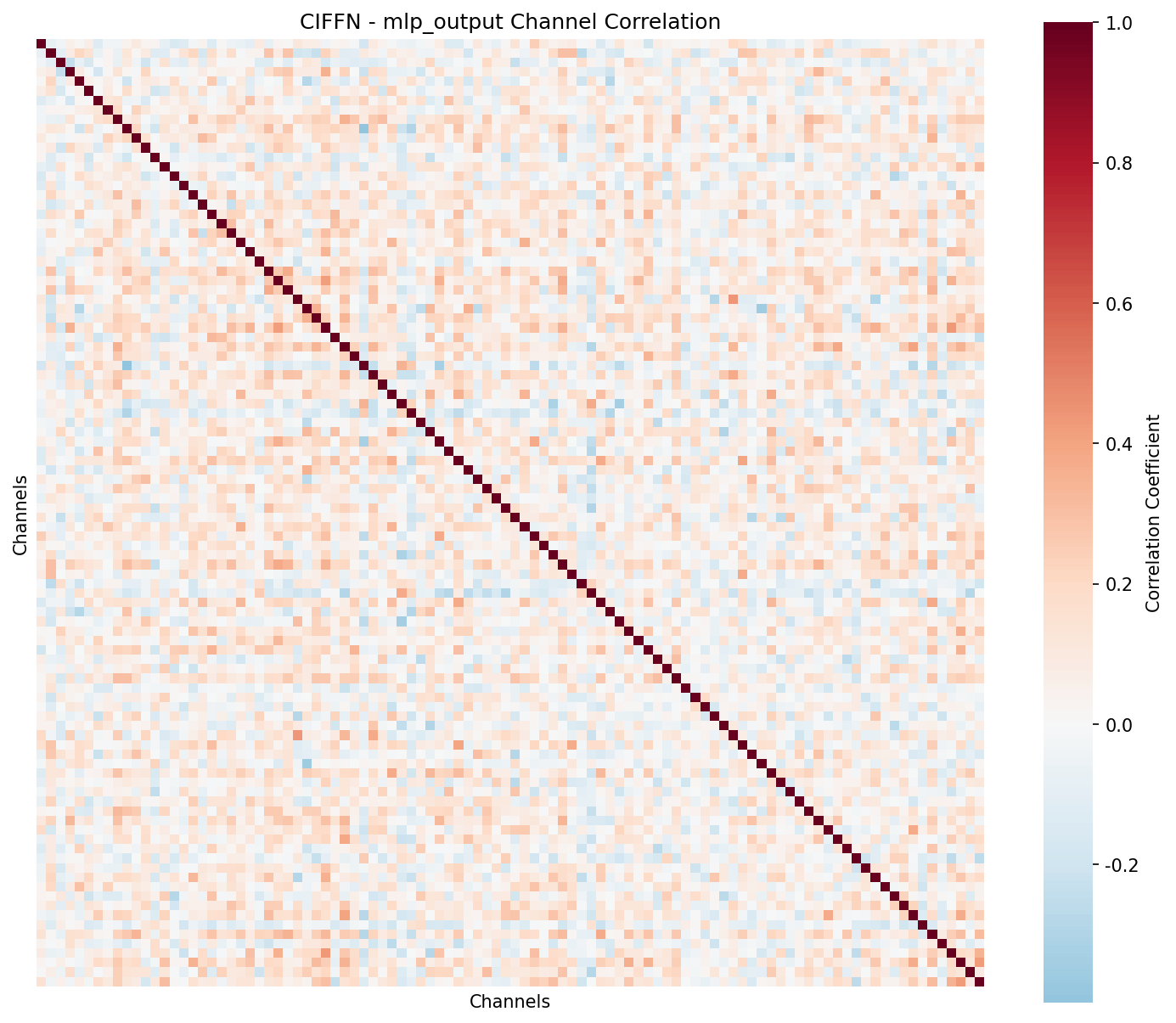}
		}
		\caption{Comparison of Channel correlation heatmap.}
		\vspace{10pt}
		\label{fig:fig5}
		\end{figure*}

	In this section, the channel correlation results among FFN, DWFFN and CIFFN will be investigated, exploring the inter-channel enhancement effect of CIFFN. To be specific, 3 models with FFN/DWFFN/CIFFN respectively are pre-trained on Tiny ImageNet. Tiny ImageNet, a sub-set of ImageNet, is a challenging benchmark for image classification. It has 200 different categories, each of which has 500 images for training, 50 for validation and 50 for testing. In this test, the intermediate features of the 3rd stage of our model (as shown in Fig. \ref{fig:fig1}) are captured in the forward propagation process to analyze the channel correlation. We randomly select 100 channels for correlation analysis through down-sampling, where we calculate the correlation matrices between different channels, and generate heatmaps and corresponding statistical information. This experiment is carried out on a laptop with AMD Ryzen 7 5800H processor and NVIDIA RTX 3060.
	As shown in Fig. \ref{fig:fig4}, the horizontal axis of the histogram shows the correlation between different channels (ranging from -1 to 1, the larger the abs mean, the more relevant they are, 0 denotes irrelevant). The vertical axis shows the number of relevant channels in each interval (i.e. frequency). Furthermore, Fig. \ref{fig:fig5} displays the heatmaps of inter-channel correlations produced by FFN, DWFFN and CIFFN respectively (both horizontal and vertical axis are selected channels). Red/Orange colors indicate higher relevance, while blue indicates less relevant, white denotes irrelevant. Based on Fig. X apparently CIFFN has lower Abs Mean and Standard deviation (Std), indicating that CIFFN has lower channel redundancy than FFN. As shown in Fig. \ref{fig:fig5}, the overall response of FFN exhibits warm colors in certain areas, reflecting the high correlation and redundancy between its channels. DWFFN, on the other hand, exhibits weak correlations by its evenly distributed cold colors, indicating strong independence among its channels (i.e. a lack of interaction). By contrast, CIFFN makes warmer colors than DWFFN, which are more evenly distributed than FFN. This phenomenon illustrates that CIFFN maintains highly correlated channel groups and necessary feature diversity while reducing the overall redundancy.
		
	\subsubsection{The role of the conv-stem}\label{backbone}
	\textcolor{red}{As shown in Table \ref{tab:tab8}, incorporating a convolutional stem improves SAEViT-T’s accuracy from 74.89\% to 76.28\% with only a minimal increase in parameters (0.06M). This improvement proves that convolutional stems better preserve low-level features such as edges and textures, which are often lost in patch-based embeddings. The conv-stem thus serves as a lightweight yet effective feature enhancer, which is particularly beneficial for extraction of fine-grained details.} 
	\begin{table}[h]
		\setlength{\tabcolsep}{3pt}
		\caption{The influence of conv-stem on SAEViT-T.}
		\centering
		\begin{tabular*}{\linewidth}{@{\extracolsep\fill}ccc@{\extracolsep\fill}} 
			\toprule
			Model Architecture&	Params(M)	&Accuracy(\%) \\
			\midrule
			Without conv-stem&	5.98	&74.89\\
			Using conv-stem	&6.04&	\textbf{76.28}\\
			\bottomrule
		\end{tabular*}
		\label{tab:tab8}
	\end{table}
	
	\section{Conclusion}
	In this paper, we propose SAEViT, a hybrid ViT based architecture for various vision tasks. The core components include Sparsely Aggregated Attention and channel-interactive Feed-Forward Network. By fusing the local prior capability of the convolution blocks with ViT's global modeling strengths, SAEViT significantly enhances the performances while maintaining relatively low computational overhead. Extensive experiments have been conducted on 3 different vision tasks to validate its effectiveness, where the advantage in the accuracy-speed tradeoff has been highlighted against other methods. Additional ablation experiments are also carried out to verify the function of the core components. 
	
	Although effective, SAEViT has some limitations under special scenarios. Firstly, the SAA module performs global static downsampling based on token redundancy, but for areas with rich details or small targets, it may cause excessive compressed information and therefore results in lower recognition accuracy for fine-grained features. Meanwhile, its fixed sampling rate also limits its adaptability to high-resolution inputs. Secondly, although CIFFN enhances inter-channel feature interaction, it still induces certain computational and memory overhead due to the depth-wise separable convolution and re-assignment operations, which may affect deployment efficiency in resource constrained environments. 
	
	\textcolor{red}{In the future, we will be focusing on 2 research directions. First, we will investigate a dynamic sparse attention mechanism that adaptively adjusts the sampling rate based on image content. For instance, a lightweight prediction module could dynamically allocate sampling rates according to local texture complexity (i.e. assign a smaller sr to texture-rich regions to preserve details and a larger sr for smooth areas to enhance computational efficiency). Second, we will conduct extensive evaluation of real-device deployment and energy efficiency, which aims to test the practical FLOPs on certain edge devices. To be specific, we plan to measure single-batch inference latency, peak memory usage, and energy consumption (in Joules per inference) across a diverse set of hardware platforms. A comparative analysis will be performed against state-of-the-art lightweight models. Key evaluation metrics will include latency (ms), frames per second (FPS), power consumption (J/inference), as well as Top-1 accuracy and mAP after quantization. These experiments will provide critical insights into the practical applicability of SAEViT in real-world edge computing scenarios. We hope our research outcome could pave the way for future researchers in this direction.}
	% Use the style of numbering in square brackets.
	% If nothing is used, default style will be taken.
	%\begin{enumerate}[a)]
	%\item 
	%\item 
	%\item 
	%\end{enumerate}  
	
	%\begin{itemize}
	%\item 
	%\item 
	%\item 
	%\end{itemize}  
	
	%\begin{description}
	%\item[]
	%\item[] 
	%\item[] 
	%\end{description}  
	
	% Figure
	% \begin{figure}[<options>]
		% 	\centering
		% 		\includegraphics[<options>]{}
		% 	  \caption{}\label{fig1}
		% \end{figure}

	% \begin{table}[<options>]
		% \begin{tabular*}{\tblwidth}{@{}LL@{}}
			% \toprule
			% \midrule
			%  & \\
			%  & \\
			%  & \\
			%  & \\
			% \bottomrule
			% \end{tabular*}
		% \end{table}
	
	% Uncomment and use as the case may be
	%\begin{theorem} 
	%\end{theorem}
	
	% Uncomment and use as the case may be
	%\begin{lemma} 
	%\end{lemma}
	

	
	\bibliographystyle{cas-model2-names}
	
	\bibliography{cas-refs}
	
	% % Here goes the biography details.
	
	% % Here goes the biography details.
	
\end{document}